\pdfoutput=1

\documentclass[11pt]{article}

\usepackage{acl}

\usepackage{times}
\usepackage{latexsym}
\usepackage{graphicx}
\usepackage[T1]{fontenc}

\usepackage[utf8]{inputenc}
\usepackage{algorithm}
\usepackage{algorithmic}
\usepackage{multirow}
\usepackage{amsfonts}
\usepackage{amsmath, mathtools}
\usepackage{microtype}

\title{Hyperbolic Relevance Matching for Neural Keyphrase Extraction}

\author{Mingyang Song,  Yi Feng and Liping Jing\thanks{\; Corresponding author.}\\
	Beijing Key Lab of Traffic Data Analysis and Mining \\
	Beijing Jiaotong University, China \\
	\texttt{mingyang.song@bjtu.edu.cn}}

\begin{document}
\maketitle
\begin{abstract}
Keyphrase extraction is a fundamental task in natural language processing that aims to extract a set of phrases with important information from a source document. Identifying important keyphrases is the central component of keyphrase extraction, and its main challenge is learning to represent information comprehensively and discriminate importance accurately. In this paper, to address the above issues, we design a new hyperbolic matching model (HyperMatch) to explore keyphrase extraction in hyperbolic space. Concretely, to represent information comprehensively, HyperMatch first takes advantage of the hidden representations in the middle layers of RoBERTa and integrates them as the word embeddings via an adaptive mixing layer to capture the hierarchical syntactic and semantic structures. Then, considering the latent structure information hidden in natural languages, HyperMatch embeds candidate phrases and documents in the same hyperbolic space via a hyperbolic phrase encoder and a hyperbolic document encoder. To discriminate importance accurately, HyperMatch estimates the importance of each candidate phrase by explicitly modeling the phrase-document relevance via the Poincaré distance and optimizes the whole model by minimizing the hyperbolic margin-based triplet loss. Extensive experiments are conducted on six benchmark datasets and demonstrate that HyperMatch outperforms the recent state-of-the-art baselines.


\end{abstract}

\section{Introduction}

Keyphrase Extraction (KE) aims to extract a set of phrases related to the main points discussed in the source document, a fundamental task in Natural Language Processing (NLP). Because of their succinct and accurate expression, keyphrase extraction is helpful for a variety of applications such as information retrieval \cite{KimKCOPS13} and text summarization \cite{LiuPLL09}. 

Typically, most existing keyphrase extraction models mainly include two procedures: candidate keyphrase extraction and keyphrase importance estimation. Specifically, the former extracts candidate phrases from the document via some heuristics \cite{heuristic3, heuristic_liu_a, heuristic5, heuristic1, heuristic_liu_b}, and the latter determines which candidate phrases are keyphrases via unsupervised or supervised models \cite{textrank, xiong19, baseline, song}. The keyphrase importance estimation procedure usually plays a more critical role than the candidate keyphrase extraction procedure in the supervised setting.

\begin{figure}
	\centering
	\includegraphics[scale=0.39]{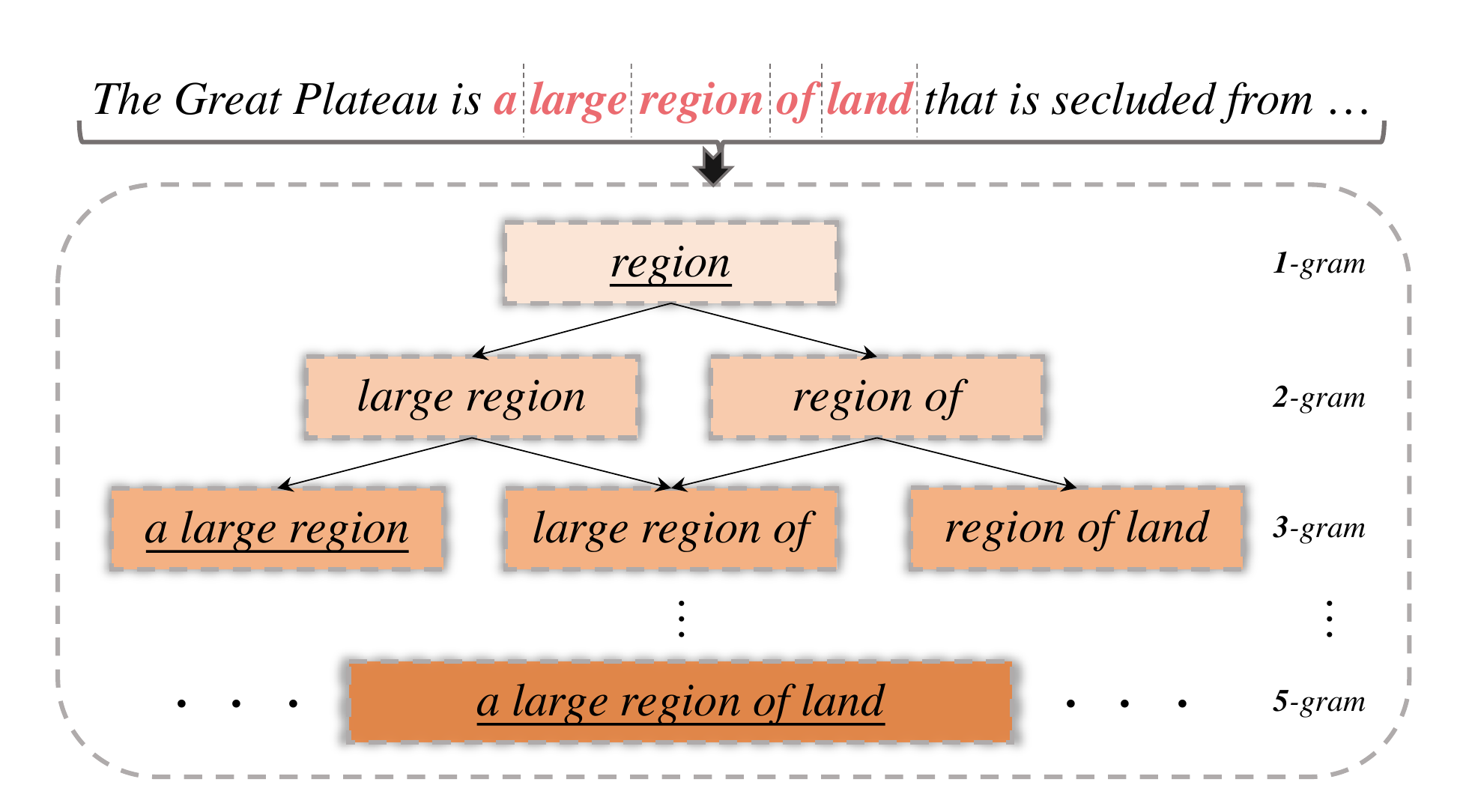}
	\caption{Sample partial of the document in \textit{OpenKP} dataset. For ease of presentation, we assume “a large region of land” is a 5-gram candidate phrase as an example in the document.}
	\label{example}
\end{figure}

In the supervised neural keyphrase extraction models, the keyphrase importance estimation procedure can be subdivided into information representation and importance discrimination. Specifically, the information representation part focuses on modeling the encoding procedure, and the importance discrimination part focuses on measuring the important scores of candidate phrases. To represent information comprehensively, recent keyphrase extraction studies have been proposed to build better representations via Bi-LSTM \cite{catseq17}, GCNs \cite{gcn2019, gcn2020}, and the pre-trained language models (e.g., ELMo \cite{xiong19}, BERT, and RoBERTa \cite{bert2020, baseline}). To discriminate the importance of candidate phrases precisely, most existing supervised keyphrase extraction models \cite{baseline, span2020, song} estimate and rank the importance of candidate phrases to extract keyphrases by using different approaches, such as classification and ranking models.

Although the existing keyphrase extraction models mentioned above have achieved significant performance, the keyphrase extraction task still needs improvement. Among them, there are the following two main issues. The first issue lies in the information representation. Typically, candidate phrases often exhibit the inherent hierarchical structures ingrained with complex syntactic and semantic information \cite{text_hyperbolic, ZhouLZ20}. In general, the longer phrases contain more complex structures. (as shown in Figure~\ref{example}, the phrase "a large region of land" has more complex inherent structures than "region" or "a large region". Similarly, the phrase "a large region" is more complex than "region"). Besides the phrases, since linguistic ontologies are intrinsic hierarchies \cite{text_hyperbolic}, the conceptual relations between phrases and their corresponding document can also form the hierarchical structures. Therefore, the hierarchical structures need to be considered when representing both phrases and documents and estimating the phrase-document relevance. However, it is difficult to capture such structural information even with infinite dimensions in the Euclidean space \cite{LinialLR95}. The second issue lies in distinguishing the importance of phrases. Keyphrases are typically used to retrieve and index their corresponding document, so they should be highly related to the main points of the source document \cite{2014survey}. However, most existing supervised keyphrase extraction methods ignore explicitly modeling the relevance between candidate keyphrases and their corresponding document, resulting in biased keyphrase extraction.

Motivated by the above issues, in this paper, we explore the potential of hyperbolic space for the keyphrase extraction task and propose a new hyperbolic relevance matching model (HyperMatch) for supervised neural keyphrase extraction. Firstly, to capture hierarchical syntactic and semantic structure information, HyperMatch integrates the hidden representations in all the intermediate layers of RoBERTa to collect the adaptive contextualized word embeddings via an adaptive mixing layer based on the self-attention mechanism. And then, considering the hierarchical structure hidden in the natural language content, HyperMatch represents both phrases and documents in the same hyperbolic space via a hyperbolic phrase encoder and a hyperbolic document encoder. Meanwhile, we adopt the Poincaré distance to calculate the phrase-document relevance by considering the latent hierarchical structures between candidate keyphrases and the document. In this setting, the keyphrase extraction task can be regarded as a matching problem and effectively implemented by minimizing a hyperbolic margin-based triplet loss. To the best of our knowledge, we are the first work to explore the supervised keyphrase extraction in hyperbolic space. Experiments on six benchmark datasets demonstrate that HyperMatch outperforms the state-of-the-art keyphrase extraction baselines. 

\begin{figure*}
	\centering
	\includegraphics[scale=0.49]{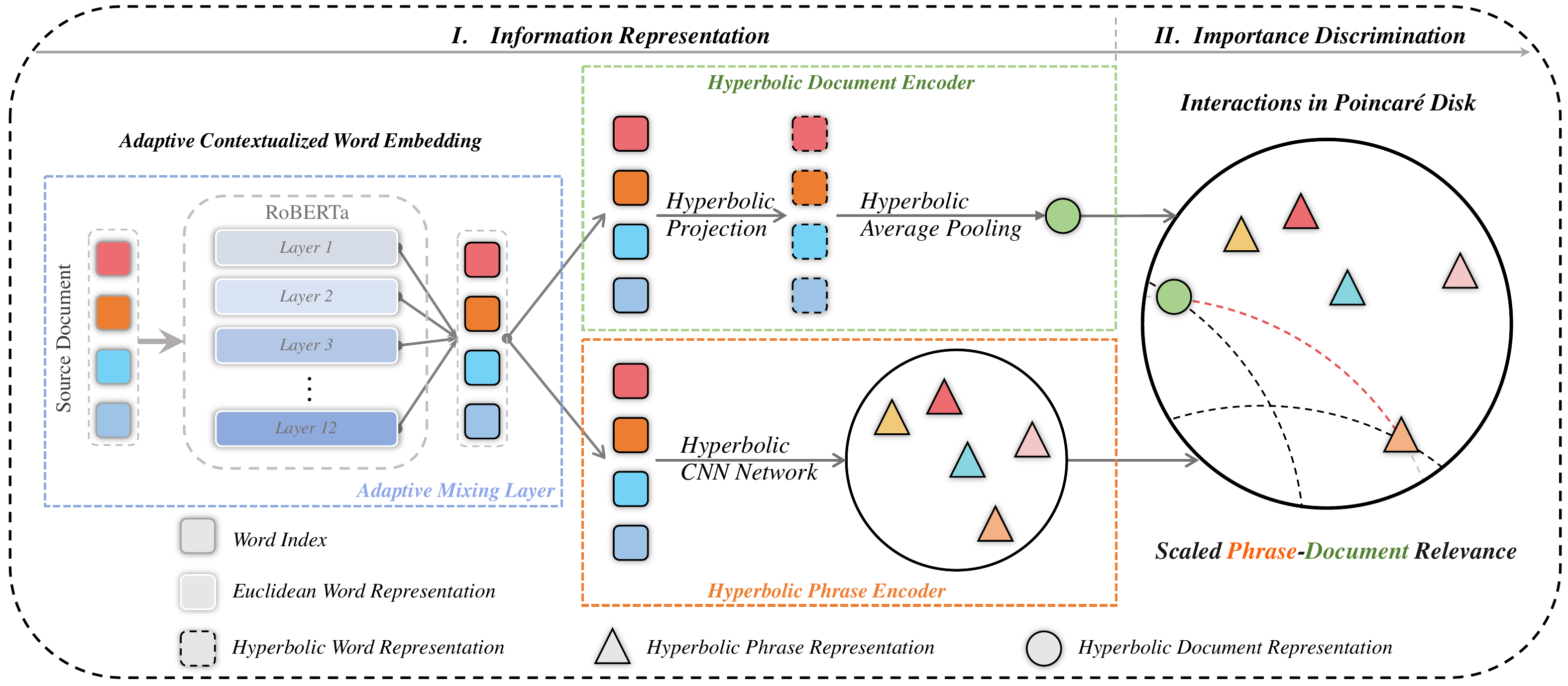}
	\caption{Framework of the hyperbolic relevance matching model (HyperMatch). }
	\label{model}
\end{figure*}

\section{Preliminaries}

Hyperbolic space is an important concept in hyperbolic geometry, which is considered as a special case in the Riemannian geometry \cite{2011Riemannian}. Before presenting our model, this section briefly introduces the basic information of hyperbolic space.

In a traditional sense, hyperbolic spaces are not vector spaces; one cannot use standard operations such as summation, multiplication, etc. To remedy this problem, one can utilize the formalism of M{\"o}bius gyrovector spaces allowing the generalization of many standard operations to the hyperbolic spaces \cite{hyperimage}.
Similarly to the previous work \cite{NickelK17, Ganea18, TifreaBG19}, we adopt the {Poincaré ball} and use an additional hyper-parameter $c$ which modifies the curvature of Poincaré ball; it is then defined as $\mathbb{D}^n_c = \{\mathbf{x}\in \mathbb{R}^n : c\Vert\mathbf{x}\Vert^2 <1, c \geq0\}$. The corresponding conformal factor now takes the form $\lambda_{\mathbf{x}}^c:= \frac{2}{1-c\Vert \mathbf{x}\Vert^2}$. In practice, the choice of $c$ allows one to balance the hyperbolic and the euclidean geometries, which is made precise by noting that when $c\rightarrow0$, all the formulas discussed below take their usual Euclidean form.  
In the following, we restate the definitions of fundamental mathematical operations for the generalized Poincaré ball model  \cite{Ganea18}. Next, we give the details of the closed-form formulas of several M{\"o}bius operations.

\noindent{{\textbf{M{\"o}bius Addition.}}}
For a pair $\mathbf{x},\mathbf{y}\in \mathbb{D}^n_c$, the {M{\"o}bius addition} is defined as,
\begin{equation}
	\small
	\mathbf{x} \oplus_c \mathbf{y} = \frac{(1+2c\langle\mathbf{x}, \mathbf{y}\rangle+ c\Vert\mathbf{y}\Vert^2)\mathbf{x}+(1-c\Vert\mathbf{x}\Vert^2)\mathbf{y}}{1+2c\langle\mathbf{x}, \mathbf{y}\rangle+c^2\Vert\mathbf{x}\Vert^2\Vert\mathbf{y}\Vert^2}.
\end{equation}

\noindent{{\textbf{M{\"o}bius  Matrix-vector Multiplication.}}}
For a linear map $\mathbf{M}: \mathbb{R}^n \rightarrow \mathbb{R}^m$ and $\forall\mathbf{x}\in \mathbb{D}^n_c$, if $\mathbf{Mx}\neq0$, 
then the {M{\"o}bius matrix-vector multiplication} is defined as,

	
\begin{equation}
	\small
	\mathbf{M} \otimes_c \mathbf{x} = (\frac{1}{\sqrt{c}})\text{tanh}(\frac{\Vert\mathbf{Mx}\Vert}{\Vert\mathbf{x}\Vert}\text{tanh}^{-1}(\Vert\sqrt{c}\mathbf{x}\Vert))\frac{\mathbf{Mx}}{\Vert\mathbf{Mx}\Vert}, 
\end{equation}
where $\mathbf{M} \otimes_c \mathbf{x}=0$ if $\mathbf{Mx}=0$.

\noindent{{\textbf{Poincaré Distance.}}}
The induced distance function is defined as,
\begin{equation}\label{distance}
	\small
	d_c(\mathbf{x},\mathbf{y}) = \frac{2}{\sqrt{c}}\text{arctanh}(\sqrt{c}\Vert-\mathbf{x}\oplus_c\mathbf{y}\Vert).
\end{equation}
Note that with $c = 1$ one recovers the geodesic distance, while with $c\rightarrow0$ we obtain the Euclidean distance $\text{lim}_{c\rightarrow0}d_c(\mathbf{x},\mathbf{y}) = 2\Vert \mathbf{x}-\mathbf{y} \Vert$. 

\noindent{{\textbf{Exponential and Logarithmic Maps.}}}
To perform operations in hyperbolic space, one first needs to define a mapping function from $\mathbb{R}^n$ to $\mathbb{D}^n_c$ to map the euclidean vectors to the hyperbolic space. Let $T_{\mathbf{x}}\mathbb{D}_c^n$ denote the tangent space of $\mathbb{D}_c^n$ at $\mathbf{x}$. The exponential map $\text{exp}^c_{\mathbf{x}}(\mathbf{\cdot}): T_{\mathbf{x}}\mathbb{D}_c^n \rightarrow \mathbb{D}_c^n$ for $\mathbf{v} \neq 0$ is defined as:
\begin{equation}
	\small
	\text{exp}^c_{\mathbf{x}}(\mathbf{v}) = \mathbf{x} \oplus_c (\text{tanh}(\sqrt{c}\frac{\lambda_{\mathbf{x}}^c\Vert\mathbf{v}\Vert}{2})\frac{\mathbf{v}}{\sqrt{c}\Vert\mathbf{v}\Vert}).
\end{equation}

As the inverse of $\text{exp}^c_{\mathbf{x}}(\mathbf{\cdot})$, the logarithmic map $\text{log}^c_{\mathbf{x}}(\mathbf{\cdot}): \mathbb{D}_c^n \rightarrow T_x\mathbb{D}_c^n$ for $\mathbf{y} \neq \mathbf{x}$ is defined as:
\begin{equation}
	\small
	\text{log}^c_{\mathbf{x}}(\mathbf{y}) = \frac{2}{\sqrt{c}\lambda_{\mathbf{x}}^c}\text{tanh}^{-1}(\sqrt{c}\Vert-\mathbf{x}  \oplus_c \mathbf{y} \Vert) \frac{-\mathbf{x}  \oplus_c \mathbf{y}}{\Vert -\mathbf{x}  \oplus_c \mathbf{y} \Vert}
\end{equation}

\noindent{{\textbf{Hyperbolic Averaging Pooling.}}}
The average pooling, as an important operation common in natural language processing, is averaging of feature vectors. In the euclidean setting, this operation takes the following form:
\begin{equation}
	\small
	\text{AP}(\mathbf{x}_1, ..., \mathbf{x}_i, ..., \mathbf{x}_M) = \frac{1}{M}\sum^M_{i=1}\mathbf{x}_i.
\end{equation}
Extension of this operation to hyperbolic spaces is called the Einstein Midpoint and takes the most simple form in Klein coordinates:
\begin{equation}\label{ap}
	\small
	\text{HyperAP}(\mathbf{x}_1, ..., \mathbf{x}_i, ..., \mathbf{x}_M) = \sum_{i=1}^{M}\gamma_i\mathbf{x}_i/ \sum_{i=1}^{M}\gamma_i,
\end{equation}
where $\gamma_i = \frac{1}{\sqrt{1-c\Vert\mathbf{x_i}\Vert^2}}$ is the Lorentz factor.
Recent work \cite{hyperimage} demonstrates that the Klein model is supported on the same space as the Poincaré ball; however, the same point has different coordinate representations in these models. Let $\mathbf{x}_{\mathbb{D}}$ and $\mathbf{x}_{\mathbb{K}}$ denote the coordinates of the same point in the Poincaré and Klein models correspondingly. Then the following transition formulas hold.

\begin{equation}\label{k}
	\small
	\mathbf{x}_{\mathbb{D}} = \frac{\mathbf{x}_{\mathbb{K}}}{1+\sqrt{1-c\Vert\mathbf{x_{\mathbb{K}}}\Vert^2}},
\end{equation}
\begin{equation}\label{p}
	\small
	\mathbf{x}_{\mathbb{K}} = \frac{2\mathbf{x}_{\mathbb{D}}}{1+c\Vert\mathbf{x_{\mathbb{D}}}\Vert^2}.
\end{equation}
Therefore, given points in the Poincaré ball, we can first map them to the Klein model via Eq.(\ref{p}), compute the average using Eq.(\ref{ap}), and then move it back to the Poincaré model via Eq.(\ref{k}).

\section{HyperMatch}
Given a document $\mathcal{D}=\{w_1, ..., w_i, ..., w_M\}$, the candidate phrases are first extracted from the source document by the n-gram rules, where $M$ indicates the max length of the input document. Then, to determine which candidate phrases are keyphrases, we design a new hyperbolic relevance matching model (HyperMatch), which mainly consists of two components: information representation and importance discrimination. Figure~\ref{model} illustrates the overall framework of HyperMatch.


\subsection{Information Representation}
Information representation is one of the essential parts of keyphrase importance estimation, which needs to represent information comprehensively. To capture rich syntactic and semantic information, HyperMatch first embeds words by the pre-trained language model RoBERTa with the adaptive mixing layer. Then, phrases and documents are embedded in the same hyperbolic space by a hyperbolic phrase encoder and a hyperbolic document encoder. In the following subsections, the information representation procedure will be described in detail.

\subsubsection{Contextualized Word Encoder}
Pre-trained language models \cite{elmo, bert,roberta} have emerged as a critical technology for achieving impressive gains in natural language tasks. These models extend the idea of word embeddings by learning contextualized text representations from large-scale corpora using a language modeling objective. Thus, recent keyphrase extraction methods \cite{xiong19, baseline, 2020sota, span2020} represent words / documents by the last intermediate layer of pre-trained language models. 

However, various probing tasks \cite{structure2019,2020specical} are proposed to discover linguistic properties learned in contextualized word embeddings, which demonstrates that different intermediate layers in pre-trained language models contain different linguistic properties or information. Specifically, each layer has specific specializations, so combining features from different layers may be more beneficial than selecting the last one based on the best overall performance.

Motivated by the phenomenon above, we propose a new adaptive mixing layer to combine all intermediate layers of RoBERTa \cite{roberta} to obtain word representations. Firstly, each word in the source document $\mathcal{D}$ is represented by all the intermediate layers in RoBERTa, which is encoded to a sequence of vector $\mathbf{H}=\{\mathbf{h}_1, ..., \mathbf{h}_i, ..., \mathbf{h}_M\}$ as follows,
\begin{equation} 
\mathbf{H}= \text{RoBERTa} \normalsize \{\mathbf{w}_1, ..., \mathbf{w}_i, ..., \mathbf{w}_M\}.
\end{equation}
Specially, $\mathbf{h}_i \in \mathbb{R}^{L*d_r}$ indicates the $i$-th contextualized word embedding of $\mathbf{w}_i$, where $L$ and $d_r$ are set to $12$ and $768$.
Then, the self-attention mechanism is adopted to aggregate the multi-layer representations of each word from RoBERTa as follows:
\begin{align}
	\alpha_{i} &= \text{softmax}(\mathbf{V}_a\mathbf{h}_{i}),\\
	\mathbf{\hat{h}}_{i} &= \mathbf{W}_a \alpha_{i}\mathbf{h}_{i},
\end{align}
where $\mathbf{V}_a\in\mathbb{R}^{d_r}$ and $\mathbf{W}_a\in\mathbb{R}^{d_r*d_r}$ denote the learnable weights. Here, $\alpha_{i}\in\mathbb{R}^{L}$ represents the adaptive mixing weights of the proposed adaptive mixing layer in HyperMatch. In this case, each word in the source document $\mathcal{D}$ is transferred to a sequence of vector $\mathbf{\hat{H}}=\{\mathbf{\hat{h}}_1, ..., \mathbf{\hat{h}}_i, ..., \mathbf{\hat{h}}_M\}$.
The adaptive mixing layer allows our model to obtain more comprehensive word embeddings, capturing more meaningful features (e.g., surface, syntactic, and semantic).

\subsubsection{Hyperbolic Phrase Encoder}
Phrases often exhibit inherent hierarchies ingrained with complex syntactic and semantic information \cite{hypertext2021}. Therefore, representing information requires sufficiently encoding semantic and syntactic information, especially for the latent hierarchical structures hidden in the natural languages. Recent studies \cite{baseline, xiong19} typically obtain phrase representations in Euclidean space, which makes it difficult to learn representations with such latent structural information even with infinite dimensions in Euclidean space \cite{LinialLR95}. On the contrary, hyperbolic spaces are non-Euclidean geometric spaces that can naturally capture the latent hierarchical structures \cite{Sarkar11,desa2018}. 

Lately, the use of hyperbolic space in NLP \cite{DhingraSNDD18, TifreaBG19, NickelK17} is motivated by the ubiquity of hierarchies (e.g., the latent hierarchical structures in phrases, sentences, and documents) in NLP tasks. Therefore, in this paper, we propose to embed phrases in hyperbolic space. Concretely, the phrase representation of the $i$-th $n$-gram $c_i^n$ is computed as follows,
\begin{equation}
	\mathbf{\hat{h}}_i^n = \text{CNN}^n (\mathbf{\hat{h}}_{i:i+n}),
\end{equation}
where $\mathbf{\hat{h}}_i^n \in \mathbb{R}^{d_h}$ represents the $i$-th $n$-gram representation, $n\in[1,N]$ indicates the length of n-grams, and $N$ is the maximum length of n-grams. Each $n$-gram has its own set of convolution filters $\text{CNN}^n$ with window size $n$ and stride $1$. 

To capture the latent hierarchies of phrases, we map phrases representation to the Poincaré ball using the exponential map, 
\begin{equation}
	\mathbf{\tilde{h}}_i^n =  \text{exp}^c_{\mathbf{0}}(\mathbf{\hat{h}}_i^n),
\end{equation}
where $\mathbf{\tilde{h}}_i^n$ is the $i$-th $n$-gram candidate phrase representation in the hyperbolic space. By mapping the representations of candidate phrases into the hyperbolic space, it is possible to implicitly capture the latent hierarchical structure of candidate phrases during the training procedure.

\subsubsection{Hyperbolic Document Encoder}
When using the source document as the query to match keyphrases, the representation of the document should cover its main points (important information). Meanwhile, documents are usually long text sequences with richer semantic and syntactic information than candidate phrases. Many current BERT-based methods \cite{span2020,matchsum} in NLP obtain documents representation by using the first output token (the [CLS] token) of the pre-trained language models. 
However, recent studies \cite{reimers2019sentencebert, LiZHWYL20} demonstrate that in many NLP tasks, documents representation obtained by the average pooling of words representation is better than the [CLS] token. 

Motivated by the above issues, we use the average pooling, a simple and effective operation, to encode documents. To further consider the latent hierarchical structures of documents, we map word representations and transfer the average pooling operation to the hyperbolic space.
In this case, we first map word representations to the hyperbolic space via the exponential map as follows:
\begin{equation}
	\mathbf{\tilde{H}} =  \{\mathbf{\tilde{h}}_1, ..., \mathbf{\tilde{h}}_i, ...,\mathbf{\tilde{h}}_M\} = \text{exp}^c_{\mathbf{0}}(\mathbf{\hat{H}}\mathbf{W}_h),
\end{equation}
where $\mathbf{W}_h\in\mathbb{R}^{d_r*d_h}$ maps the original BERT embedding space to the tangent space of the origin of the Poincaré ball. Then $\text{exp}_{\mathbf{0}}(\cdot)$ maps the tangent space inside the Poincaré ball. Next, we use the hyperbolic averaging pooling to encode the source document as follows: 
\begin{equation}
	\mathbf{\tilde{h}} =\text{HyperAP}(\{\mathbf{\tilde{h}}_1, ..., \mathbf{\tilde{h}}_i, ...,\mathbf{\tilde{h}}_M\}),
\end{equation}
where $\mathbf{\tilde{h}}\in \mathbb{R}^{d_h}$ indicates the hyperbolic document representation (called Einstein Midpoint pooling vectors in the Poincaré ball \cite{Gulcehre19}). The hyperbolic average pooling emphasizes semantically specific words that usually contain more information but occur less frequently than general ones. It should be noted that points near the boundary of the Poincaré ball get larger weights in the Einstein Midpoint formula, which may be more representative content in the source document \cite{DhingraSNDD18, hypertext2021}.

\subsection{Importance Discrimination}
Importance discrimination is one of the primary parts of the keyphrase importance estimation procedure, which estimates the important scores of candidate phrases accurately to extract keyphrases. To reach this goal, we first calculate the scaled phrase-document relevance between candidate keyphrases and their corresponding document via the Poincaré distance as the important score of each candidate keyphrase. Then, the whole model is optimized by the hyperbolic margin-based triplet loss to extract keyphrases accurately.

\subsubsection{Scaled Phrase-Document Relevance}
Besides the intrinsic hierarchies of linguistic ontologies, the conceptual relations between candidate phrases and their corresponding document can also form hierarchical structures.
Once the document representation $\mathbf{\tilde{h}}$ and phrase representations $\mathbf{\tilde{h}}_i^n$ are obtained, it is expected that the phrases and their corresponding document embedded close to each other based on their geodesic distance\footnote{Note that cosine similarity \cite{WangHF17} is not appropriate to be the metric since there does not exist a clear hyperbolic inner-product for the Poincaré ball \cite{TifreaBG19}, so the Poincaré distance is more suitable.} if they are highly relevant. Specifically, the scaled phrase-document relevance of the $i$-th $n$-gram representation $c_i^n$ can be computed as follows:
\begin{equation}\label{s}
	\textit{S}(c_i^n, \mathcal{D}) = - \frac{\lambda (d_c (\mathbf{\tilde{h}}_i^n, \mathbf{\tilde{h}}))^2}{\sqrt{d_{h}}} + (1-\lambda)f_c(\mathbf{\tilde{h}}_i^n),
\end{equation}
where $\textit{S}(\cdot)$ indicates the scaled phrase-document relevance. Here, $d_c$ indicates the Poincaré distance, which is introduced in Eq.(\ref{distance}). Here, $f_c$ indicates the linear transformation in hyperbolic space. Specifically, for Eq.~\ref{s}, the first term models the phrase-document relevance explicitly, and the second term models the phrase-document relevance implicitly. Estimating the phrase-document relevance via the Poincaré distance in hyperbolic space allows HyperMatch to model the latent hierarchical structures between candidate phrases and their document, accurately estimating the importance of candidate keyphrases. In addition, we find that increasing the dimension $d_h$ of representations will increase the value of the phrase-document relevance, causing the optimization collapse of our model. To counteract this effect, we scale the phrase-document relevance by $\frac{1}{\sqrt{d_{h}}}$.

\subsubsection{Margin-based Triplet Loss}

To select phrases with higher importances, we adopt the margin-based triplet loss in our model and optimize for margin separation in hyperbolic space.
Therefore, we first treat the candidate keyphrases in the document that are labelled as keyphrases, in the positive set $\mathbf{P^+}$, and the others to the negative set $\mathbf{P^-}$, to obtain the matching labels. Then, the loss function is calculated as follows:
\begin{equation}
	\mathcal{L} =\text{max} (0, \frac{\delta}{\sqrt{d_{h}}} -\textit{S}(p^+, \mathcal{D}) \\+ \textit{S}(p^-, \mathcal{D})),
\end{equation}
where $\delta$ indicates the margin. It enforces HyperMatch to sort the candidate keyphrases $p^+$ ahead of $p^-$ within their corresponding document. Through this training objective, our model will tend to extract the keyphrases, which are more relevant to the source document.

\section{Experimental Settings}

\subsection{Benchmark Datasets}

\noindent Six benchmark keyphrase datasets are used in our experiments, which contain \textit{OpenKP} \cite{xiong19}, \textit{KP20k} \cite{catseq17}, \textit{Inspec} \cite{Inspec}, \textit{Krapivin} \cite{Krapivin}, \textit{Nus} \cite{Nus}, and \textit{SemEval} \cite{SemEval}). We follow the previous work \cite{baseline} to preprocess each dataset with the same procedure.




\subsection{Implementation Details}
Implementation details of HyperMatch are summarized in Table~\ref{parameter}.
The maximum document length is 512 tokens due to RoBERTa limitations \cite{roberta} and documents are zero-padded or truncated to this length.
Our model was implemented in Pytorch 1.8\footnote{https://pytorch.org/} \cite{pytorch} using the hugging face reimplementation of RoBERTa\footnote{https://huggingface.co/transformers/index.html} \cite{transformer_pytorch} and was trained on eight NVIDIA RTX A4000 GPUs to achieve the best performance.


\begin{table}[!t]
	\small
	\centering
	\renewcommand\tabcolsep{4pt}
	\renewcommand\arraystretch{1.3}
	\begin{tabular}{cc}
		\hline \hline
		\textbf{Hyperparameter} & \textbf{Dimension or Value} \\ 
		\hline
		RoBERTa Embedding  $(\mathbb{R}^{d_c})$ & 768 \\ 
		Hyperbolic Rank $(\mathbb{R}^{d_h})$ & 768 \\ 
		Max Sequence Length & 512 \\ 
		Maximum Phrase Length $(N)$ & 5 \\
		\hline
		$c$ & $1$ \\
		$\lambda$ & $0.5$ \\
		$\delta$ & 1.0 \\
		Optimizer & AdamW \\
		Batch Size & $72$ \\
		Learning Rate & $5\times10^{-5}$ \\
		Warm-Up Proportion & $10\%$ \\ 
		\hline \hline
	\end{tabular}
	\caption{Parameters used for training HyperMatch.}
	\label{parameter}
\end{table}
\subsection{Evaluation Metrics}

For the keyphrase extraction task, the performance of the existing models is typically evaluated by comparing the top-$K$ predicted keyphrases with the target keyphrases (the ground-truth labels).
The evaluation cutoff $K$ can be a fixed number (e.g., F1@5 compares the top-$5$ keyphrases predicted by the model with the ground-truth to compute an F1 score).
Following the previous work \cite{catseq17, baseline, song}, we adopt macro-averaged recall and F-measure (F1) as evaluation metrics, and $K$ is set to be 1, 3, 5, and 10.
In the evaluation, we apply Porter Stemmer\footnote{https://tartarus.org/martin/PorterStemmer/} to both the target keyphrases and the extracted keyphrases when determining the exact match of keyphrases.

\begin{table*}[!htb]
	\small
	\centering
	\renewcommand\tabcolsep{7.7pt}
	\renewcommand\arraystretch{1.5}
	\begin{tabular}{l|ccc|ccc|ccc}
		\hline\hline
		\multirow{2}{*}{\normalsize \textbf{{Model}}} & \multicolumn{9}{c}{\textbf{\textit{OpenKP}}} \\\cline{2-10} 
		& P@1 & {P@3} & P@5  & R@1 & {R@3} & R@5   & F1@1 & \textbf{F1@3} & F1@5 \\ \hline
		\multicolumn{10}{l}{{Unsupervised Keyphrase Extraction Models}}\\\hline
		\multicolumn{1}{l|}{{TFIDF}} 
		& {28.3} & {18.4} & {13.7} & {15.0} & {28.4} & {34.7} & {19.6}$^\dagger$ & {22.3}$^\dagger$ & {19.6}$^\dagger$ \\
		
		\multicolumn{1}{l|}{{TextRank}} 
		& 7.7 & 6.2 & 5.5 & 4.1 & 9.8 & 14.2 & 5.4$^\dagger$ & 7.6$^\dagger$ & 7.9$^\dagger$ \\
		\hline
		
		
		\multicolumn{10}{l}{{Supervised Keyphrase Extraction via Classification Models}}\\\hline
		\multicolumn{1}{l|}{{{BERT-Chunking-KPE}}} 
		& 51.1 & 30.6 & 22.5 & 27.1 & 46.4 & 55.8 & 34.0  & 35.6  & 31.1 \\
		\multicolumn{1}{l|}{{{SpanBERT-Chunking-KPE}}} 
		& 52.3 & 32.1 & 23.5 & 27.8 & 48.6 & 58.1 & 34.8  & 37.2  & 32.4 \\
		\multicolumn{1}{l|}{{{RoBERTa-Chunking-KPE}}} 
		& 53.3 & 32.2 & 23.5 & 28.3 & 48.6 & 58.1 & 35.5  & 37.3  & 32.4 \\
		\hline
		\multicolumn{10}{l}{{Supervised Keyphrase Extraction via Ranking Models}}\\\hline
		\multicolumn{1}{l|}{{{BERT-Ranking-KPE}}} 
		& 51.3 & 32.3 & 23.5 & 27.3 & 48.9 & 58.2 & 34.2  & 37.4  & 32.5 \\
		
		\multicolumn{1}{l|}{{{SpanBERT-Ranking-KPE}}} 
		& 53.0 & 32.7 & 24.0 & 28.4 & 49.7 & 59.3 & 35.5  & 38.0  & 33.1 \\
		
		\multicolumn{1}{l|}{{{RoBERTa-Ranking-KPE}}} 
		& 53.8 & 33.7 & 24.4 & 29.0 & 50.9 & 60.4 & 36.1  & 39.0  & 33.7 \\
		\hline
		\multicolumn{1}{l|}{{\textbf{HyperMatch}}} 
		& \textbf{54.7} & \textbf{33.9} & \textbf{24.7} & \textbf{29.5} & \textbf{51.5} & \textbf{61.2} & \textbf{36.4}  & \textbf{39.4}  & \textbf{33.7}\\
		
		\hline\hline
	\end{tabular}
	\caption{Model performance on the \textit{OpenKP} dataset. The best results of our model are highlighted in bold. F1@3 is the main evaluation metric (marked in bold) for this dataset \cite{xiong19, 2020sota}. $^\dagger$ denotes these results are not included in the original paper and are estimated with Precision and Recall score. The results of the baselines are reported in their corresponding papers.}
	\label{openkp}
\end{table*}

\subsection{Baselines}

We compare two kinds of solid baselines to give a comprehensive evaluation of the performance of HyperMatch: unsupervised keyphrase extraction models (e.g., TextRank \cite{textrank} and TFIDF \cite{tfidf}) and supervised keyphrase extraction models (e.g., classification and ranking models based variants of BERT \cite{baseline}).
Noticeably, HyperMatch extracts keyphrases without using additional features on the \textit{OpenKP} dataset. Therefore, for the sake of fairness, we do not compare with the methods \cite{xiong19, 2020sota} which use additional features to extract keyphrases. 

In addition, this paper mainly focuses on exploring keyphrase extraction in hyperbolic space via a matching framework (similar to the ranking model). Hence, the compared baselines we mainly choose are keyphrase extraction methods based on the classification and ranking models rather than some existing studies based on integration models \cite{2021select,2021uni} or multi-task learning \cite{song}.

\section{Results and Analysis}
In this section, we test the performance of HyperMatch on six widely-used benchmark keyphrase extraction datasets (\textit{OpenKP}, \textit{KP20k}, \textit{Inspec}, \textit{Krapivin}, \textit{Nus}, and \textit{Semeval}) from three facets. The first one demonstrates its superiority by comparing HyperMatch with the recent baselines in terms of several metrics. The second one is to verify the effect of each component via ablation tests. The last one is to test the sensitivity of the hyperbolic margin-based triplet loss with different margins.
\subsection{Performance Comparison}
The experimental results are given in Table~\ref{openkp} and Table~\ref{kp20k}. Overall, HyperMatch outperforms the recent BERT-based keyphrase extraction models (the results are reported in their own articles) in most cases. Concretely, on the \textit{OpenKP} and \textit{KP20k} datasets, HyperMatch achieves better results than the best ranking models RoBERTa-Ranking-KPE. The main reason for this result may be that learning representation in hyperbolic space can capture more latent hierarchical structures than the euclidean space. Meanwhile, compared with the results on the other four zero-shot datasets (\textit{Inspec}, \textit{Krapivin}, \textit{Nus}, and \textit{Semeval}) in Table~\ref{kp20k}, it can be seen that HyperMatch outperforms both unsupervised and supervised baselines. We consider that the main reason is the scaled phrase-document relevance \textit{explicitly }models a strong connection between phrases and their corresponding document via the Poincaré distance, obtaining more robust performance even in different datasets.


\begin{table*}[!htb]
	\small
	\centering
	\renewcommand\tabcolsep{5pt}
	\renewcommand\arraystretch{1.5}
	\begin{tabular}{l|cc|cc|cc|cc|cc}
		\hline\hline
		\multirow{2}{*}{\normalsize \textbf{\textit{Model}}} & \multicolumn{2}{c|}{\textit{\textbf{Inspec}}} & \multicolumn{2}{c|}{\textit{\textbf{Krapivin}}}& \multicolumn{2}{c|}{\textit{\textbf{Nus}}}& \multicolumn{2}{c|}{\textit{\textbf{SemEval}}}& \multicolumn{2}{c}{\textit{\textbf{KP20k}}}\\ \cline{2-11} 
		& F1@5 & F1@10 & F1@5 & F1@10 & F1@5 & F1@10& F1@5 & F1@10& F1@5 & F1@10\\ \hline

		\multicolumn{1}{l|}{{TFIDF}} 
		& 22.3 & 30.4 & 11.3 & 14.3 & 13.9 & 18.1 & 12.0 & 18.4 & 10.5 & 13.0 \\
		\multicolumn{1}{l|}{{TextRank}} 
		& 22.9 & 27.5 & 17.2 & 14.7 & 19.5 & 19.0 & 17.2 & 18.1 & 18.0 & 15.0 \\
		\hline
		
		
		
		
		\multicolumn{1}{l|}{{RoBERTa-Ranking-KPE$^\dagger$}} 
		& 28.1 & 29.1 & 29.9 & 23.7 & 44.6 & 37.7 & 35.4 & 32.6 & 41.4 & 34.2 \\

		\multicolumn{1}{l|}{{\textbf{HyperMatch}}}
		& \textbf{30.4} & \textbf{32.2} & \textbf{32.8} & \textbf{26.3} & \textbf{45.8} & \textbf{41.3} & \textbf{35.7} & \textbf{36.8} & \textbf{41.6} & \textbf{34.3} \\
		\hline\hline
	\end{tabular}
	\caption{Results of keyphrase extraction on five benchmark keyphrase datasets. F1 scores on the top $5$ and $10$ keyphrases are reported. $^\dagger$ indicates that these results are evaluated via the code which is provided by its corresponding paper. The best results are highlighted in bold.}
	\label{kp20k}
\end{table*}
\subsection{Ablation Study}

\begin{table}[!htb]
	\small
	\centering
	\renewcommand\tabcolsep{5pt}
	\renewcommand\arraystretch{1.5}
	\begin{tabular}{l|ccc}
		\hline\hline
		\multirow{2}{*}{\normalsize \textbf{{Model}}} & \multicolumn{3}{c}{\textit{\textbf{OpenKP}}} \\\cline{2-4} 
		&  F1@1 & \textbf{F1@3} & F1@5   \\ \hline
		
		\multicolumn{1}{l|}{{{{\textbf{HyperMatch}}}}}
		& \textbf{36.4}  & \textbf{39.4} & \textbf{33.7} \\ 
		
		\multicolumn{1}{l|}{{{EuclideanMatch}}} 
		& 36.1  & 38.5  & 33.4 \\
		
		\multicolumn{1}{l|}{{{HyperMatch w/o Relevance}}} 
		& 36.1 & 38.9  & 33.6 \\
		
		\multicolumn{1}{l|}{{{{{HyperMatch w/o AML}}}}}
		& 36.3  & 38.7  & 33.5  \\

		\hline\hline
	\end{tabular}
	\caption{Ablation tests on the \textit{OpenKP} dataset. The best results are highlighted in bold. F1@3 is the main evaluation metric (marked in bold) for this dataset.}
	\label{ablation}
\end{table}

In this section, we report on several ablation experiments to analyze the effect of different components. The ablation experiment on the \textit{OpenKP} dataset is shown in Table~\ref{ablation}.

To measure the effectiveness of hyperbolic space for the keyphrase extraction task, we compare it with the same model in the euclidean space and use the euclidean distance to explicitly model the phrase-document relevance. 
As shown in Table~\ref{ablation}, HyperMatch outperforms EuclideanMatch, which shows that using the hyperbolic space can capture the latent hierarchical structures more effectively than the euclidean space. 

To verify the effectiveness of the adaptive mixing layer, we propose a model HyperMatch w/o AML, which indicates HyperMatch without using the adaptive mixing layer module and only uses the last intermediate layer of RoBERTa to embed phrases and documents.
As shown in Table~\ref{ablation}, the performance of our model HyperMatch without using the adaptive mixing layer drops in all evaluation metrics. These results demonstrate that combining all the intermediate layers of RoBERTa may capture more helpful information (e.g., surface, syntactic, and semantic) for obtaining candidate phrases and documents representations.

Unlike our model, most recent keyphrase extraction methods (e.g., RoBERTa-Ranking-KPE) implicitly model relevance between candidate phrases and their corresponding document by a linear transformation layer as the phrase-document relevance.
Therefore, to verify the effectiveness of explicitly modeling the phrase-document relevance, we built the HyperMatch w/o Relevance, which only implicitly computes the phrase-document relevance by the hyperbolic linear transformation layer \cite{Ganea18}.
The results of HyperMatch w/o Relevance show a drop in all evaluation metrics, indicating that explicitly considering the relevance between phrases and the document is essential for estimating the importance of candidate phrases in the keyphrase extraction task.

\begin{figure}[!htb]
	\centering
	\includegraphics[scale=0.45]{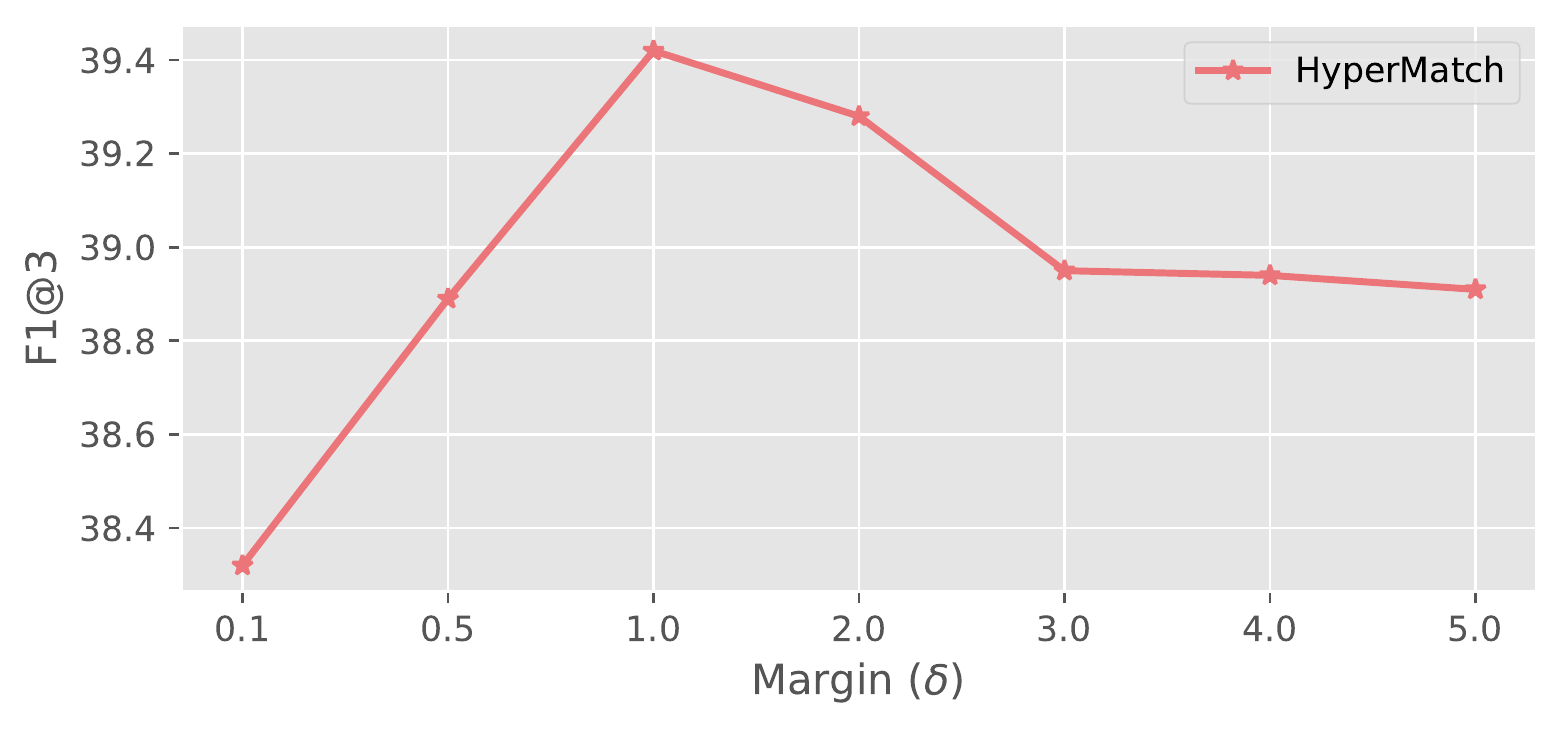}
	\caption{Performance of HyperMatch with different margins ($\delta$) of the margin-based triplet loss on the \textit{OpenKP} dataset.}
	\label{margin_rank}
\end{figure}
\subsection{Sensitivity of Hyperparameters}
In this section, we verify the sensitivity of HyperMatch with different margins ($\delta$) of the hyperbolic margin-based triplet loss.
For keyphrase extraction methods equipped with the margin-based triplet loss, the margin design significantly impacts the final result, where a poor margin usually causes performance degradation. Therefore, we verify the effects of different margins on HyperMatch in Figure~\ref{margin_rank}. We can see that HyperMatch achieves the best results when $\delta=1$.

\section{Related Work}
This section briefly describes the related work from two fields: keyphrase extraction and hyperbolic deep learning.
\subsection{Keyphrase Extraction}
Most existing KE models are based on the two-stage extraction framework, which consists of two main procedures: candidate keyphrase extraction and keyphrase importance estimation. Candidate keyphrase extraction extracts a set of candidate phrases from the document by some heuristics (e.g., essential n-gram-based phrases \cite{hulth2004,medelyan2009,xiong19,baseline,2020sota}). Keyphrase importance estimation first represents candidate phrases and documents by the pre-trained language models \cite{bert, roberta} and then estimates the phrase-document relevance implicitly as the importance scores. Finally, the candidate phrases are ranked by their importance scores, which can be learned by either unsupervised \cite{textrank, heuristic_liu_b} or supervised \cite{xiong19,baseline,span2020} ranking approaches.

Different from the existing KE models, we map phrases and documents representations from the euclidean space to the same hyperbolic space to capture the latent hierarchical structures. Next, we adopt the Poincaré distance to explicitly model the phrase-document relevance as the important score of each candidate phrase. Finally, the hyperbolic margin-based triplet loss is used to optimize the whole model. To the best of our knowledge, we are the first study to explore supervised keyphrase extraction in hyperbolic space.

\subsection{Hyperbolic Deep Learning}
Recent studies on representation learning \cite{NickelK17,TifreaBG19,MathieuLMTT19} demonstrate that hyperbolic space is more suitable for embedding symbolic data with hierarchies than the Euclidean space since the tree-like properties \cite{hamann2018tree} of the hyperbolic space make it efficient to learn hierarchical representations with low distortion \cite{desa2018,Sarkar11}. As linguistic ontologies are innately hierarchies, hierarchies are ubiquitous in natural language \cite{text_hyperbolic}. Some recent studies show the superiority of hyperbolic space for many natural language processing tasks \cite{Gulcehre19, hypertext2021}. \citet{chen2021probing} demonstrate that mapping contextualized word embeddings (i.e., BERT-based embeddings) to the hyperbolic space can capture richer hierarchical structure information than the euclidean space when encoding natural language text. Inspired by the above methods, we transfer the embeddings obtained by the pre-trained language models to hyperbolic space for extracting keyphrases.

\section{Conclusions and Future Work}

A new hyperbolic relevance matching model HyperMatch is proposed to map candidate phrases and documents representations into the hyperbolic space and model the relevance between candidate phrases and the document via the Poincaré distance. Specifically, HyperMatch first combines the intermediate layers of RoBERTa via the adaptive mixing layer for capturing richer syntactic and semantic information. Then, phrases and documents are encoded in the same hyperbolic space to capture the latent hierarchical structures. Next, the phrase-document relevance is estimated explicitly via the Poincaré distance as the importance scores of all the candidate keyphrases. Finally, we adopt the hyperbolic margin-based triplet loss to optimize the whole model for extracting keyphrases.

In this paper, we explore keyphrase extraction in hyperbolic space and implicitly model the latent hierarchical structures hidden in natural languages when representing candidate keyphrases and documents. In the future, it will be interesting to introduce external knowledge (e.g., WordNet) to explicitly model the latent hierarchical structures when representing candidate keyphrases and documents. In addition, our code is publicly available to facilitate other research\footnote{https://github.com/MySong7NLPer/HyperMatch}. 

\section{Acknowledgments}
This work was supported in part by the National Key Research and Development Program of China under Grant 2020AAA0106800; the National Science Foundation of China under Grant 61822601 and 61773050; the Beijing Natural Science Foundation under Grant Z180006; the Fundamental Research Funds for the Central Universities (2019JBZ110).

\bibliography{anthology}
\bibliographystyle{acl_natbib}

\end{document}